\documentclass[lettersize,journal]{IEEEtran}
\usepackage{amsmath,amsfonts}
\usepackage{algorithmic}
\usepackage{array}
\usepackage[caption=false,font=normalsize,labelfont=sf,textfont=sf]{subfig}
\usepackage{textcomp}
\usepackage{stfloats}
\usepackage{url}
\usepackage{verbatim}
\usepackage{graphicx}
\usepackage[sort]{natbib}
\def\BibTeX{{\rm B\kern-.05em{\sc i\kern-.025em b}\kern-.08em
    T\kern-.1667em\lower.7ex\hbox{E}\kern-.125emX}}
\usepackage{balance}
\begin{document}
\title{Embodied Neuro-symbolic AI:\\Making Robot Behavior Auditable, Safe, and Scalable}
\author{Ruwan Wickramarachchi\textsuperscript{1}, Jonathan Francis\textsuperscript{1,2}, Alessandro Oltramari\textsuperscript{1}
\thanks{\textsuperscript{1} Bosch Center for Artificial Intelligence, Pittsburgh, PA, USA. \textsuperscript{2} Carnegie Mellon University, Pittsburgh, PA, USA.\\
\texttt{\{ruwan.wickramarachchi, alessandro.oltramari, jon.francis\}@us.bosch.com}}}
%


\maketitle
\begin{abstract}
Current Vision-Language-Action (VLA) end-to-end systems have driven major progress in robot learning, yet they still face persistent barriers to dependable real-world deployment. In practice, these models often exhibit brittle generalization under distribution shift (i.e., novel objects, layouts, lighting, or contact conditions), limited physical and causal reasoning, and inconsistent fine-grained motor control. Critically, many VLA pipelines offer limited ways to express and audit what a robot must not do, making it difficult to provide assurance in safety-critical settings. These limitations constrain scalability, maintainability, and trust when robots move beyond controlled demonstrations. We propose Embodied Neuro-Symbolic (NeSy) AI: a deployment-oriented paradigm that integrates learned perception, planning and control with explicit, machine-readable background knowledge and formal constraints. By combining knowledge-guided perception, constraint-aware symbolic planning, and hybrid learning, Embodied NeSy AI enables robots to reason about their environment, adapt to new tasks, and provide auditable, machine-checkable compliance with safety and operational requirements, thus supporting scalable, trustworthy robot deployment.
\end{abstract}


\section{Introduction}

The next generation of reliable, auditable, and scalable robot intelligence will require moving beyond the limitations of current end-to-end neural systems. While deep learning and VLA models have enabled robots to learn complex sensorimotor skills from demonstration and large-scale data, they remain fundamentally limited in their ability to generalize, reason about physical and causal structure, and provide verifiable guarantees for safety and compliance. These limitations are especially acute in open-world, safety-critical, or rapidly changing environments, where implicit statistical regularities are insufficient for robust deployment.

Embodied NeSy AI addresses these challenges by integrating neural perception and control with explicit, machine-readable knowledge and formal constraints. Rather than relying solely on implicit statistical regularities, Embodied NeSy makes key domain structure—task semantics, operational procedures, hardware/product metadata, and safety requirements—explicit and actionable inside the decision loop. This knowledge is represented using ontologies, knowledge graphs, and constraint specifications, capturing site and use-case configurations, robot and tool capabilities/limits (e.g., payload, workspace envelopes, calibration, speed/force bounds), subject-matter expertise, standard operating procedures, and formal safety rules (e.g., restricted zones, proximity constraints, permitted contact regimes).

Central to this framework are three tightly integrated components: (1) knowledge-guided semantic perception that grounds symbolic predicates in sensor data while tracking uncertainty; (2) constraint-aware symbolic planning that generates machine-checkable action sequences satisfying task goals and formal safety requirements; and (3) hybrid learning that updates both neural skills and symbolic knowledge from experience under safety envelopes. By combining these elements, Embodied NeSy AI enables robots to reason about their environment, adapt to new tasks, and provide auditable, machine-checkable compliance with safety and operational requirements—supporting compositional adaptation, continual learning, and clearer governance for trustworthy, large-scale robot deployment in diverse domains.

The need for such integration becomes evident when considering the shortcomings of single-paradigm approaches. Purely neural embodied systems lack interpretability, offer no guarantees of constraint satisfaction, generalize poorly to novel task compositions, and require extensive retraining for new scenarios. Classical symbolic approaches, in contrast, can face challenges in grounding symbols in noisy sensor data, learning and adapting from experience, scaling to complex state spaces, and representing uncertainty. Embodied NeSy AI bridges these weaknesses by combining data-driven learning with structured reasoning.

\section{Embodied NeSy AI Framework: Components and Integration}

\begin{figure}
    \centering
    \includegraphics[width=\linewidth]{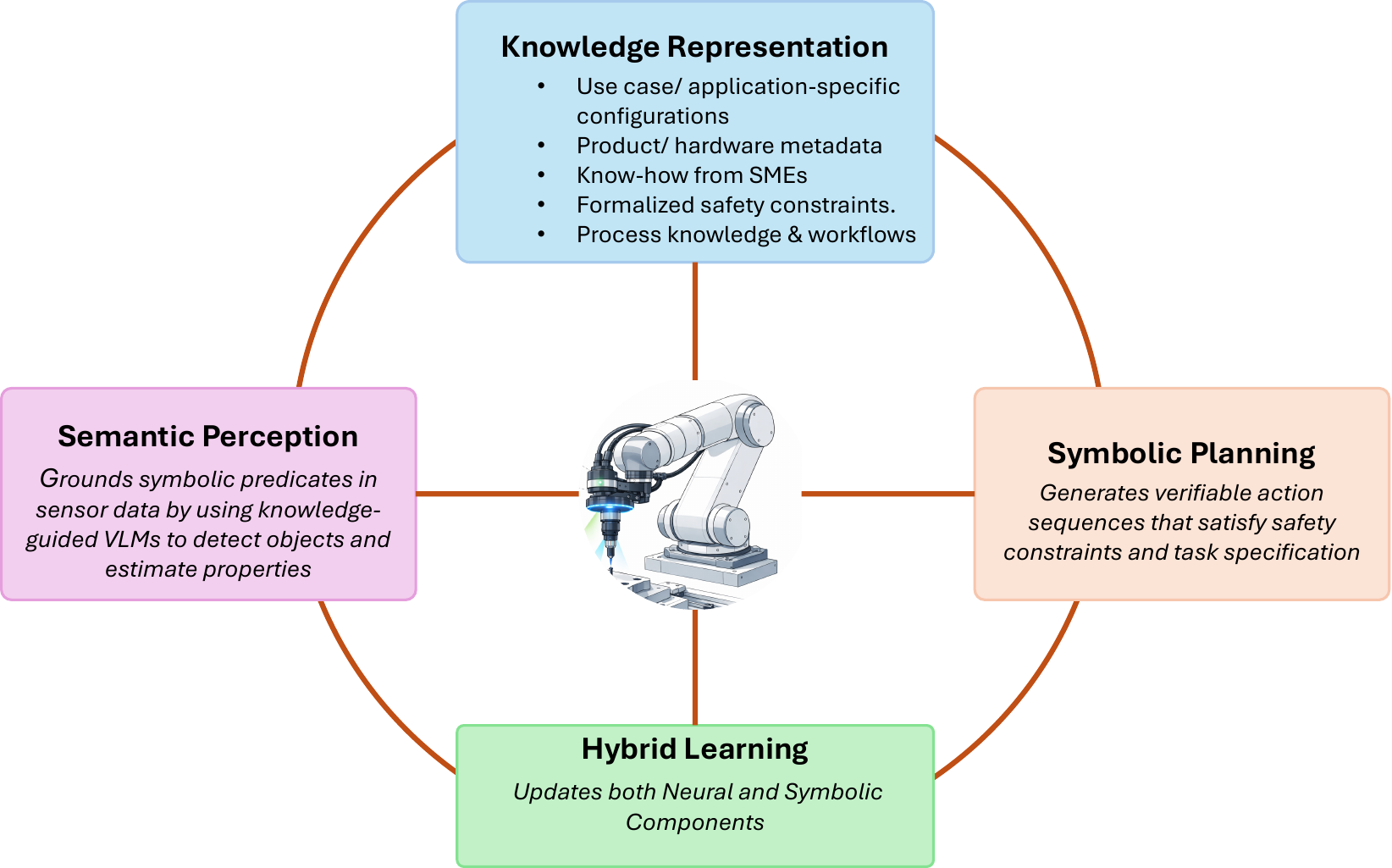}
    \caption{Main components of Embodied NeSy AI Framework}
    \vspace{-1em}

    \label{fig:embodied-nesy-framework}
\end{figure}

Embodied NeSy AI is built on three tightly integrated components (see Figure \ref{fig:embodied-nesy-framework}), each leveraging explicit knowledge representation for robust, auditable, and adaptive robot behavior. Knowledge representation is central to this framework: ontologies, knowledge graphs, and constraint specifications encode site and use-case configurations, robot and tool capabilities/limits (e.g., payload, workspace envelopes, calibration, speed/force bounds), subject-matter expertise, standard operating procedures, and formal safety rules (e.g., restricted zones, proximity constraints, permitted contact regimes). By making these elements structured and auditable, Embodied NeSy AI supports compositional adaptation, continual learning, and clearer governance during deployment. The three core components of this framework are explained below.\\

\begin{enumerate}
     \item \textbf{Semantic perception (knowledge-guided grounding with uncertainty):} Perception modules employ Vision-Language Models (VLMs) or Vision-Language-Action (VLA) systems, guided by ontologies and knowledge graphs, to ground symbolic predicates in sensor streams (e.g., object identity, affordances, zones, human proximity) and estimate task-relevant properties (e.g., pose, fragility cues, graspability), while tracking uncertainty/confidence. The output is an uncertainty-aware, queryable belief state—comprising knowledge graph triples and logical predicates—that encodes entities, properties, geometry, and relations. This structured representation enables downstream symbolic planners to validate goals, decompose tasks, and enforce constraints by querying the current state of the world, supporting robust operation under ambiguity and novelty.\\

    \item \textbf{Constraint-aware symbolic planning (specification to plans):} Symbolic planners generate only those action sequences that satisfy task goals, symbolic/task specifications, safety rules, and temporal logic constraints \emph{before} execution. This pre-execution enforcement ensures that plans are filtered for reachability, forbidden states, restricted zones, tool/process prerequisites, and correct ordering of actions. The planner checks each candidate plan against formal constraints—such as state/action bounds, temporal safety rules, and process dependencies—so that only compliant plans are produced. VLA plan priors (e.g., likely action sequences or skill policies suggested by Vision-Language-Action models) can also be incorporated to guide or bias the symbolic planner, improving efficiency and leveraging learned experience. All plans are machine-checkable with respect to the abstract model and stated assumptions, providing explicit, auditable guarantees of constraint satisfaction at the planning level. This enables robots to reason about safety, compliance, and operational requirements in a transparent and verifiable manner, reducing the risk of unsafe or invalid behavior before any action is taken.\\ 

    \item \textbf{Hybrid learning (continual updates across perception, planning, and knowledge):} Hybrid learning in Embodied NeSy AI operates at three levels: (1) \emph{Perception:} Neural modules improve grounding accuracy and adapt thresholds for symbolic predicate extraction based on feedback and uncertainty; (2) \emph{Planning:} Heuristics and plan priors are tuned from experience to improve efficiency and robustness of symbolic planners; (3) \emph{Knowledge:} The symbolic layer is continually refined through knowledge graph completion, rule learning, and rule refinement, enabling the system to expand and update its knowledge base over time. Unlike conventional systems that keep symbolic knowledge static, Embodied NeSy AI continually refines the symbolic layer via rule learning, knowledge graph completion, or program synthesis under invariant-preserving safeguards. These continual updates allow rapid adaptation to new tasks and environments without full-stack retraining, supporting scalable and robust deployment.
\end{enumerate}

\section{Related Work}

Neurosymbolic AI aims to unify subsymbolic learning with symbolic reasoning, addressing the limitations of classical symbolic approaches and the opacity of end-to-end neural policies \citep{garcezneurosymbolic2023,shethneurosymbolic2023}. We organize related work along three dimensions central to our Embodied NeSy AI framing: knowledge-guided perception and grounding, constraint-aware planning and verification, and hybrid learning and adaptation.\\ 

\noindent\textbf{Knowledge-guided perception and grounding.} Explicit knowledge representations complement neural models with structured information about entities, relations, physical properties, affordances, and domain constraints. Ontologies such as KNOWROB support action-centered representations and reasoning about objects, actions, and affordances \cite{bessler2020formal}, while knowledge graphs support hybrid modeling and neural-symbolic reasoning \cite{rudolph2024hybrid,tran2025reasoningneurosymbolicai}. Recent approaches further combine neural perception with explicit symbolic relations for tasks such as spatial reasoning and object localization \citep{jahangard2025multi}. Foundation models such as VLMs and VLAs provide powerful semantic priors but do not inherently maintain explicit, machine-queryable representations of task-relevant entities, constraints, and uncertainty. Our framing therefore emphasizes an uncertainty-aware symbolic belief state that connects learned perception, structured knowledge, and downstream reasoning.\\ 

\noindent\textbf{Constraint-aware planning, safety, and verification.} Neurosymbolic architectures increasingly combine learned models with symbolic planning and formal constraints. Modular approaches pair LLM-based reasoning with neural controllers \citep{ali2025control}, while systems such as PFEA use vision-language planning, execution feedback, and replanning for robotic manipulation \citep{ding2025pfea}. Safety-oriented approaches introduce specification-driven architectures and verification against explicit safety constraints \citep{zheng2025neurostrata,yang2025repv}, with related work addressing reasoning shortcuts and assurance in neural-symbolic systems \citep{richards2025promise,marconato2025prototypical}. These methods demonstrate the value of explicit reasoning and verification, particularly for safety-critical embodied systems. Our framing integrates these capabilities by placing constraint-aware symbolic planning directly between grounded perception and robot execution, enabling plans to be checked against task, operational, and safety constraints before execution.\\

\noindent\textbf{Hybrid learning and adaptation.} Recent work also investigates tighter integration of neural learning and symbolic reasoning. Imperative Learning couples neural modules with classical solvers and model-based optimization \citep{wang2025imperative,li2025ikap}; Neurosymbolic Imitation Learning learns symbolic abstractions for decomposing tasks and sequencing neural skills \citep{keller2025nsil}; and Symbolic Imitation Learning uses Inductive Logic Programming to induce interpretable rules \citep{sharifi2025sil}. Simulation-based approaches further support generalization through large-scale pretraining and synthetic data \citep{interndata2025}. While these approaches provide mechanisms for joint learning, abstraction, and generalization, adaptation often centers on individual components. Our framing instead treats learning as continual adaptation across perception, planning, and explicit knowledge, allowing the system to incorporate experience without requiring full-stack retraining while preserving safety-critical invariants.\\

Together, these directions provide key building blocks for embodied neurosymbolic autonomy. Our Embodied NeSy AI framework emphasizes their integration: knowledge-guided perception constructs an uncertainty-aware symbolic belief state, constraint-aware planning produces machine-checkable plans under explicit specifications, and hybrid learning continually adapts perception, planning, and knowledge. Explicit knowledge representation serves as the common interface across these components, supporting auditable, safe, and scalable robot behavior.

\section{Application Domains}

Embodied NeSy AI enables robust, interpretable, and scalable robot deployment. We highlight four representative settings where knowledge-guided perception, constraint-aware planning, and safety-bounded learning add immediate value:\\

\begin{figure*}[!ht]
\centering
\includegraphics[width=\textwidth]{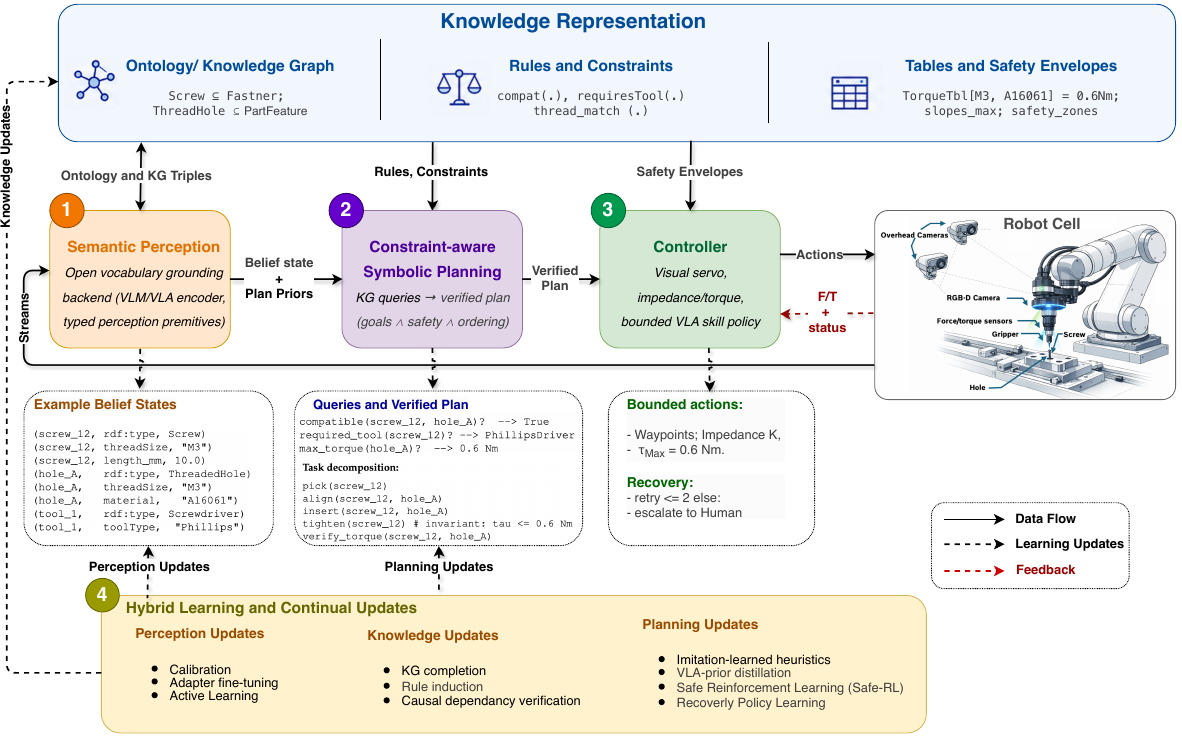}
\caption{Use-case diagram of a precision screw fastening pipeline, illustrating the interactions among Knowledge Representation, Semantic Perception, Symbolic Planning, Controller, and Hybrid Learning components.}

\label{fig:screw-fastening}
\end{figure*}

\begin{itemize}
    \item \textbf{Manufacturing Cell:} Robot performing screwing operation and being able to adapt across product variants while meeting torque and alignment limits; detect and recover from cross-threading with auditable rules. Figure~\ref{fig:screw-fastening} illustrates our running example.
    \item \textbf{Household and humanoids:} Combine natural language tasking with safe, compositional skill execution and interpretable failure modes for assistance (e.g., elderly and patient care) tasks.
    \item \textbf{Autonomy and mobility:} Plan with explicit safety envelopes in uncertain environments; verify plans against symbolic specifications before execution.
    \item \textbf{Simulation and evaluation:} Use structured knowledge and uncertainty tracking to close sim-to-real gaps and to benchmark robustness, safety, and explainability.
\end{itemize}

Beyond manufacturing, the same integration pattern applies across service robotics, mobility, and autonomy, where transparency and safety are first-class requirements.

\section{Illustrative Use Case: Precision Screw Fastening}

To demonstrate how the Embodied NeSy AI architecture can be applied to concrete use cases,  we consider a precision screw fastening task in manufacturing. This scenario highlights the technical interplay between three stages, namely semantic perception, symbolic planning, and hybrid learning. In manufacturing automation, precision screw fastening in automotive assembly requires adaptive perception, fine-grained force control, and process verification across product variants. State-of-the-art VLA models (e.g., $\pi_{0.6}$, RT-2, OpenVLA, Pelican-VL) struggle with millimeter-precision localization, torque specification compliance, and interpretable failure diagnosis. Pure symbolic approaches, meanwhile, cannot robustly process sensor data or adapt to environmental variation.

Embodied NeSy architectures address these limitations through layered integration: semantic perception provides knowledge-guided scene understanding, symbolic planning decomposes operations into verifiable phases with explicit safety constraints, and hybrid learning enables continual adaptation. Knowledge graphs encode fastener specifications, material properties, assembly sequences, expert know-how, and safety rules---supporting rapid adaptation to new product variants. This integration enables generalization, formal constraint compliance, and interpretable audit trails---capabilities essential for safety-critical deployment and large-scale automation. Figure~\ref{fig:screw-fastening} shows the use-case diagram used in this work.

\subsection{Stage 1: Semantic Perception (Knowledge-Guided Grounding)}

\textbf{Inputs:} camera stream (RGB-D wrist and overhead cameras), wrist force/torque signal, joint encoders.
\textbf{Output:} a typed belief state---knowledge-graph triples annotated with per-triple provenance and confidence---consumed by Stage~2.\\

An open-vocabulary perception backend performs detection, attribute extraction, phrase grounding, and geometry estimation against the vocabulary supplied by the ontology (see the Knowledge Representation bar of Fig.~\ref{fig:screw-fastening}). Concrete instantiations include (i) a Vision-Language Model (VLM), (ii) the perception encoder of a Vision-Language-Action (VLA) policy---only the encoder outputs are consumed here, no action head is invoked at this stage---or (iii) a library of typed perception primitives (e.g., \texttt{detect\_object}, \texttt{get\_pose}, \texttt{grasp\_score}) backed by specialized open-vocabulary grounders such as Grounded-SAM  \cite{ren2024grounded}, in the style of Code-as-Policies (CaP)~\citep{liang2023code}. The output form is identical across backends, so downstream stages are unaffected by the choice.\\

\noindent \textbf{Example VLM output (with provenance):}
\begin{itemize}
\small
    \item Detected objects: \texttt{screw\_12}~(Screw, VLM 0.93), \texttt{hole\_A}~(ThreadedHole, VLM 0.88), \texttt{tool\_1}~(Screwdriver, VLM 0.91).
    \item VLM-estimated attributes: \texttt{screw\_12.threadSize = M3} (0.70), \texttt{screw\_12.length = 10\,mm} (0.65), \texttt{tool\_1.toolType = Phillips} (0.82).
    \item Deterministic attributes from Computer-aided Design (CAD) /Manufacturing Execution System (MES), that's represented in KG, for the loaded variant: \texttt{hole\_A.threadSize = M3}, \texttt{hole\_A.material = Al6061}.
    \item Geometry via RGB-D fusion: \texttt{pose(screw\_12), pose(hole\_A)} $\in SE(3)$.
    \item Symbolic relations derived from geometry: \texttt{near(screw\_12, hole\_A)}, \texttt{aligned(screw\_12.axis, hole\_A.axis)}.\\
\end{itemize}

\noindent \textbf{Triples populated in the knowledge graph} (each carrying source and confidence, matching the ``Example belief state'' callout in Fig.~\ref{fig:screw-fastening}):
\small
\begin{verbatim}
(screw_12, rdf:type, Screw)  [0.93]
(screw_12, threadSize, "M3") [VLM, 0.70]
(screw_12, length_mm, 10.0) [VLM, 0.65]
(hole_A,   rdf:type, ThreadedHole) [VLM, 0.88]
(hole_A,   threadSize, "M3") [CAD, 1.00]
(hole_A,   material,   "Al6061") [KR, 1.00]
(tool_1,   rdf:type, Screwdriver) [VLM, 0.91]
(tool_1,   toolType,  "Phillips") [VLM, 0.82]
(screw_12, candidateTarget, hole_A) [derived]
\end{verbatim}

Uncertainty is retained on every VLM-sourced triple so downstream planning can gate low-confidence facts, while CAD/MES-sourced triples are asserted deterministically.

\subsection{Stage 2: Constraint-Aware Symbolic Planning}
\textbf{Inputs:}  the Stage-1 belief state (KG triples with provenance and confidence), the ontology/rule store, and a task goal supplied by the operator or MES (e.g., \texttt{fastened(screw\_12, hole\_A)}).\\
\textbf{Output:} a verified action sequence---together with the KG queries and invariants that certify it---ready for dispatch to the controller.\\

Given these inputs, the planner queries the ontology and knowledge graph to retrieve the metadata and safety constraints relevant to the task goal, and decomposes it into a sequence of actions that provably satisfy those constraints. Concretely, Stage 2 can be realized with a symbolic planning formalism such as Planning Domain Definition Language (PDDL) (e.g., PDDL 2.1 for durative/numeric constraints), a Hierarchical Task Network (HTN) encoding for hierarchical decomposition, or a task-and-motion planning (TAMP) framework (e.g., PDDLStream) when continuous geometric constraints from Stage 1 must be interleaved with symbolic reasoning. LLM-driven alternatives are also compatible: a code-generator such as Code-as-Policies~\citep{liang2023code} emits programs, and a graph-generator such as Graph-as-Policy~\citep{chen2026gap} emits directed computation graphs of perception, planning, and control nodes over a skill library---in both cases operating over the same KG-plus-primitives interface (KG queries, Stage-1 perception primitives, control operators), provided the emitted plans are still checked against the ontology-derived constraints before dispatch.\\

\textbf{Knowledge graph / ontology encodes:}
\begin{itemize}
\small
    \item Type hierarchy: \texttt{Screw} $\sqsubseteq$ \texttt{Fastener}, \texttt{ThreadedHole} $\sqsubseteq$ \texttt{PartFeature}.
    \item Compatibility: \texttt{ThreadedHole(M3) compatibleWith Screw(M3)}.
    \item Tool requirements: \texttt{Screw(M3, headType=Phillips) requiresTool(PhillipsDriver)}.
    \item Process rules: only \texttt{tighten} if \texttt{thread\_match(screw, hole)} holds.
    \item Safety table lookup: \texttt{TorqueTbl[threadSize, material]}, cross-thread detection via torque-slope bound $d\tau/d\theta \le s_{\max}$.
\end{itemize}

\noindent\textbf{Symbolic planner queries} (each answered by an ontology axiom or a KG lookup, as annotated):
\small
\begin{verbatim}
compatible(screw_12, hole_A)?  --> True  
required_tool(screw_12)? --> PhillipsDriver 
max_torque(hole_A)?  --> 0.6 Nm
\end{verbatim}

\noindent The torque bound $0.6$\,Nm is not a free-standing constant---it is the value returned by the deterministic KG table lookup on the material asserted from CAD in Stage~1.

\noindent\textbf{Task decomposition:}
\small
\begin{verbatim}
pick(screw_12)
align(screw_12, hole_A)
insert(screw_12, hole_A)
tighten(screw_12) # invariant: tau <= 0.6 Nm
verify_torque(screw_12, hole_A)
\end{verbatim}

All actions are checked against constraints before dispatch (e.g., $\tau$ must not exceed the queried maximum, and the tighten step aborts if the torque slope rises too quickly), so only compliant plans reach the controller.

\subsection{Stage 3: Hybrid Learning and Continual Knowledge Update}

\textbf{Inputs:} execution traces (actions, sensor logs, outcomes), episode summaries, and the current KG / neural checkpoints.
\textbf{Output:} updated encoder / skill-policy weights, KG deltas (SME-gated), and refit planner heuristics.

By \emph{hybrid} we mean a learning regime in which two update substrates run concurrently and cross-supervise each other: gradient-based updates of neural parameters (perception encoder, skill policy) and discrete updates of symbolic artifacts (KG triples, rules, table entries). Neural evidence drives symbolic updates, and symbolic constraints regularize the neural updates. In the running use case this yields three update streams---one per component of the pipeline:\\

\begin{itemize}
    \item \textbf{Perception updates:} improve grounding and recalibrate uncertainty on VLM-sourced triples using accumulated outcome feedback, keeping the neural encoder consistent with ontology axioms.
    \item \textbf{Knowledge updates:} extend and refine the KG (new triples, refined rules, recalibrated numeric tables such as \texttt{TorqueTbl[M3, Al6061]}: $0.60 \rightarrow 0.55$\,Nm after repeated cross-thread events) from observed successes and failures, gated by SME review as an invariant-preserving safeguard.
    \item \textbf{Planning updates:} tune planner heuristics and the controller's skill policy from execution traces, and learn a recovery-vs.-escalate policy from failure signatures such as torque-slope spikes.\\
\end{itemize}

At runtime, anomalies (e.g., unexpected torque spike) still trigger the symbolic recovery routine (reverse, re-align, retry, else escalate); every episode's outcome is appended to the KG and to the training buffer, closing the loop between execution and the three update streams above.

This use case illustrates how the three stages---semantic perception, symbolic planning, and hybrid learning---work together to deliver robust, auditable, and adaptive robot behavior in a real-world manufacturing scenario.

\vspace{-1em}
\section{Conclusion}

Embodied NeSy AI represents a paradigm shift for trustworthy, scalable, and auditable robot deployment. By tightly integrating knowledge-guided perception, constraint-aware symbolic planning, and hybrid learning, this approach overcomes the limitations of purely neural or symbolic systems. Explicit knowledge representation and formal constraints enable robots to reason about their environment, adapt to new tasks, and provide verifiable guarantees for safety and compliance. As research advances in neurosymbolic integration, simulation, and verification, Embodied NeSy AI is poised to deliver robust, interpretable, and continually improving autonomy across diverse real-world domains.
\vspace{-0.5em}

\bibliographystyle{unsrt}
\bibliography{references}

\end{document}